\documentclass[11pt]{article}

\usepackage[final]{acl}

\usepackage{times}
\usepackage{latexsym}

\usepackage[T1]{fontenc}

\usepackage[utf8]{inputenc}

\usepackage{microtype}

\usepackage{inconsolata}

\usepackage{graphicx}
\usepackage{amssymb}
\usepackage{amsmath}
\usepackage{subcaption}
\usepackage{booktabs}
\usepackage{multirow}

\title{Privacy-Preserving Reasoning with Knowledge-Distilled \\Parametric Retrieval Augmented Generation}

\author{
  Jinwen Chen$^{1,2}$ \quad
  Hainan Zhang$^{1,2}$\thanks{\ \ Corresponding author: \texttt{zhanghainan@buaa.edu.cn}} \quad
  Liang Pang$^3$ \\
  Yongxin Tong$^{4}$ \quad
  Haibo Zhou$^{5}$ \quad
  Yuan Zhan$^{5}$ \quad
  Wei Lin$^{5}$ \quad
  Zhiming Zheng$^{1,2}$ \\
  $^1$Beijing Advanced Innovation Center for Future Blockchain and Privacy Computing \\
  $^2$School of Artificial Intelligence, Beihang University, China \\
  $^3$Institute of Computing Technology, Chinese Academy of Sciences \\
  $^4$School of Computer Science and Engineering, Beihang University, China \\
  $^5$Meituan \\
  \texttt{\{jwkami, zhanghainan\}@buaa.edu.cn}
}

\begin{document}
\maketitle
\begin{abstract}
The current RAG system requires uploading plaintext documents to the cloud, risking private data leakage. Parametric RAG (PRAG) encodes documents as LoRA parameters within LLMs, offering a possible way to reduce exposure of raw content. However, it still faces two issues: (1) PRAG demands synthesizing QA pairs and fine-tuning LLM for each individual document to create its corresponding LoRA, leading to \textbf{unacceptable inference latency}. (2) The performance of PRAG relies solely on synthetic QA data while lacking internal alignment with standard RAG, resulting in \textbf{poor generalization} on out-of-distribution(OOD) inputs. Therefore, achieving high-efficiency parameterization while maintaining RAG-level performance remains a critical challenge for privacy-preserving reasoning. In this paper, we propose DistilledPRAG, a generalizable knowledge-distilled parametric RAG model aligned with standard RAG in document structure and parameter activation. We first synthesize QA pairs from single and multi-documents to enhance cross-document reasoning. Then, we mask the plaintext documents with a special token and translate them to LoRA via a parameter generator, maintaining the standard RAG document structure. Finally, guided by synthetic QA data, we train the parameter generator to match standard RAG's hidden states and output logits, enabling RAG-style reasoning without original documents. Experiments on four QA datasets show that DistilledPRAG outperforms baselines in accuracy and generalizes well on OOD data~\footnote{https://github.com/JWQZ/DistilledPRAG-arxiv}.
\end{abstract}

\section{Introduction}

\begin{figure}[!t]
\centering
\includegraphics[width=0.9\columnwidth]{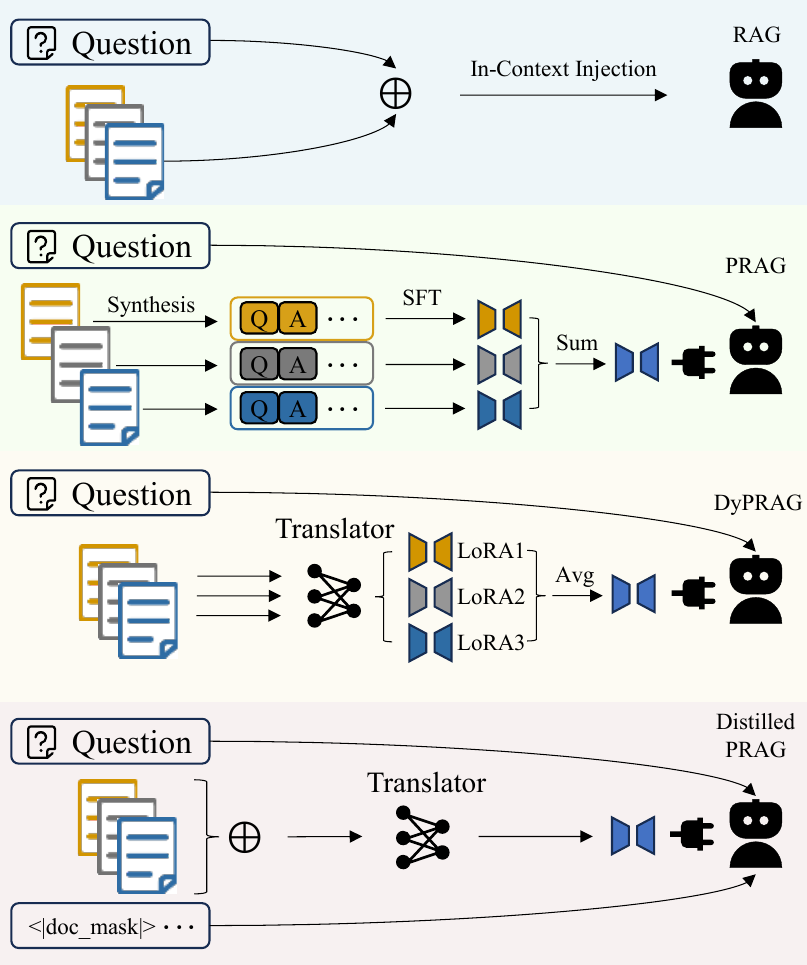} 
\caption{\label{fig:motivation} Inference Paradigms for standard RAG, PRAG, DyPRAG, and our DistilledPRAG. (1) Standard RAG inputs the plaintext documents and the question. (2) PRAG generates QA pairs per document to fine-tune LoRA adapters, and sums them to obtain document aggregated representations for LLM injection. (3) DyPRAG translates individual documents to its LoRA and averages them to achieve document aggregation for LLM injection. (4) DistilledPRAG generates cross-document LoRA from concatenated documents via a parameter generator, then takes masked documents and questions as input, similar to standard RAG. }
\end{figure}

Retrieval-Augmented Generation (RAG)~\cite{rag1,rag2,dpr} enables real-time integration of external knowledge into LLMs and is widely used across fields like finance~\cite{zhang2023enhancing}, law~\cite{louis2024interpretable}, and materials science~\cite{buehler2024generative}. However, current RAG systems require uploading local documents to the cloud, raising privacy concerns when handling sensitive information such as corporate contracts, medical records, or personal notes~\cite{zeng2024good}. Therefore, a new RAG paradigm without plaintext is necessary to support privacy-preserving reasoning.

Recently, parametric RAG (PRAG)~\cite{su2025parametricretrievalaugmentedgeneration} is proposed to encode documents into LoRA parameters~\cite{hu2022lora} and upload to a cloud server, thereby avoiding the transmission of plaintext documents. As shown in Figure~\ref{fig:motivation}, PRAG first synthesizes QA pairs for an individual document and then fine-tunes LLM with them to obtain its corresponding LoRA parameters. In inference, PRAG aggregates the LoRAs of retrieved documents to answer questions. However, the need for both data synthesis and fine-tuning per document leads to unacceptable inference latency in real-world scenarios.

To reduce this latency, DyPRAG~\cite{tan2025dynamicparametricretrievalaugmented} proposes training a dynamic parameter translator to replace data synthesis and fine-tuning at test-time. It still synthesizes QA pairs and fine-tunes LLM to obtain the corresponding LoRA like PRAG, but differs by training a linear parameter translator to map each document to its LoRA. As shown in Figure~\ref{fig:motivation}, during inference, each retrieved document is processed by the parameter translator to generate its LoRA, which is then averaged and injected into the LLM to generate the final answer.

However, both PRAG and DyPRAG rely solely on synthetic QA pairs to activate knowledge learning, lacking internal alignment with standard RAG, which may lead to \textbf{poor generalization} on out-of-distribution(OOD) inputs. This misalignment appears in two ways: (1) Document structure: training LoRA on individual documents and aggregating them disrupts cross-document reasoning. For instance, in Figure~\ref{fig:motivation}, summing or averaging LoRA1, LoRA2, and LoRA3 will miss cross-document reasoning cues, leading to incorrect answers(see in Appendix~\ref{app:cross-psg}). (2) Parameter activation: performing inference only with the query alters internal parameter activations compared to standard RAG, losing original reasoning capability. Therefore, efficient parameterization with RAG-level reasoning remains a major challenge for parametric RAG.

In this paper, we introduce DistilledPRAG, a knowledge-distilled parametric RAG model by aligning the document structure and parameter activations with a standard RAG for more robust generalization on OOD data. Specifically, we first use DeepSeek-V3 to synthesize QA pairs for individual documents and concatenated multi-documents to enhance cross-document reasoning. Then, we replace plaintext document tokens with a special token and translate them to LoRA by our LongT5-based parameter generator, forming the student model with the same document structure as standard RAG(teacher). Finally, guided by the synthetic QA data, we train the parameter generator by minimizing the differences between the student and teacher in both hidden states and output logits, enabling it to learn RAG-style reasoning without access to the original documents. In this setting, the student model inherits the teacher model's structure and activations, allowing rapid learning of standard RAG reasoning.

Experiments on four public QA datasets demonstrate that DistilledPRAG, trained only on 2WQA, outperforms baselines on three OOD datasets, validating its strong generalization capabilities. Further analysis on synthetic data and alignment functions confirms the effectiveness of our internal alignment mechanism. Our main contributions are:
\begin{itemize}
\item We identify that internal alignment between parametric RAG and standard RAG is crucial, facilitating efficient and highly generalizable document-specific parameter learning.

\item We propose a knowledge-distilled parametric RAG by internal alignment with standard RAG in terms of both document structure and activation parameters, thereby enabling robust generalization to OOD data.

\item Experiments show that DistilledPRAG delivers strong QA performance without uploading plaintext documents, and exhibits a superior ability to convert unseen documents into reasoning-capable LoRA modules.

\end{itemize}

\section{Related Work}

Recently, Parametric Retrieval-Augmented Generation (PRAG)~\cite{su2025parametricretrievalaugmentedgeneration} proposes encoding retrieved documents into model parameters (e.g., via LoRA) instead of appending them as context. This approach injects external knowledge directly into the LLMs, treating retrieval as parameter updates rather than input expansion. To obtain the document's parameters, PRAG requires synthesizing QA pairs and fine-tuning LLMs for each retrieved document, which has unacceptable inference latency in real-world testing scenarios. To address this, Dynamic PRAG (DyPRAG)~\cite{tan2025dynamicparametricretrievalaugmented} is proposed to introduce a parameter translator that maps documents to parameters at test-time, enabling document-specific parameterization without additional fine-tuning or storage overhead. DyPRAG improves flexibility but often fails to generalize across OOD inputs, limiting its ability to replace traditional RAG under privacy constraints fully. To overcome these issues, we propose a knowledge-distilled parametric RAG method by cross-document data augmentation and aligning the model’s behavior across the student and teacher models, significantly enhancing its generalization ability to OOD inputs.

\section{Backgrounds}

\begin{figure*}[!t]
\centering
\includegraphics[width=0.9\textwidth]{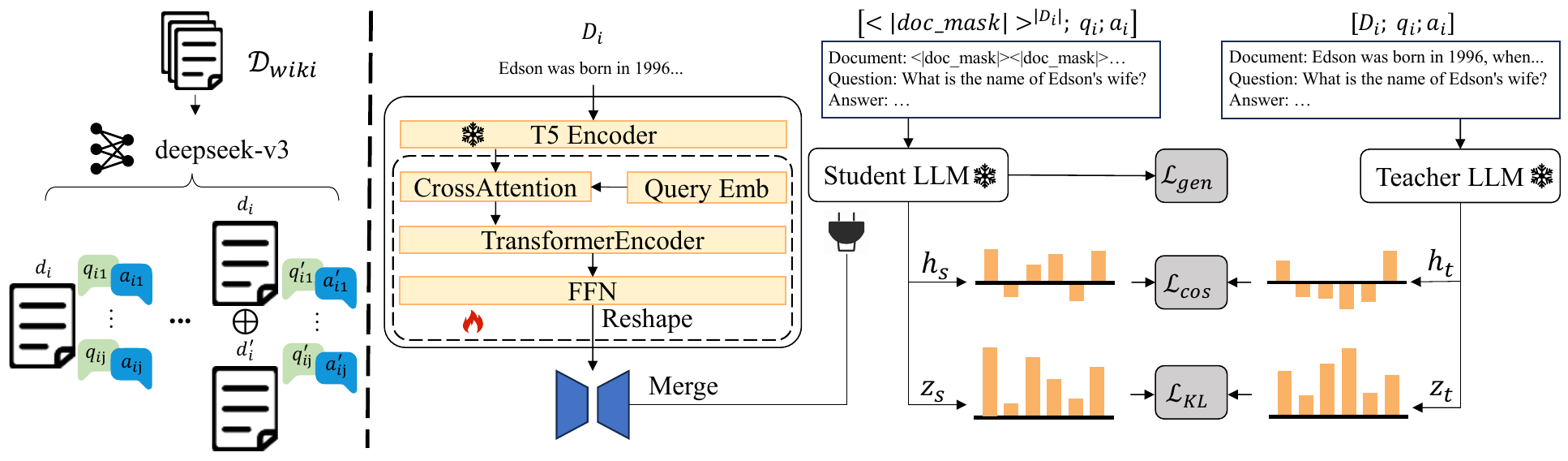}
\caption{\label{fig:main} The Architecture of DistilledPRAG Model. \textcircled{{\small 1}}Use DeepSeek-V3 to mine knowledge from a single document and augmented cross-documents by random concatenation. \textcircled{{\small 2}}Train a parameter generator to map documents to a LoRA for student LLM, enabling it to mimic a teacher RAG's reasoning by minimizing differences in hidden states and logits on synthetic data. }

\end{figure*}

Now, we will introduce the process and notions of standard RAG and PRAG. Let \( q \) denote a user query and \( D=\{d_1,\cdots,d_N\} \in \mathcal{C} \) denote the top-N documents retrieved from a large corpus \( \mathcal{C} \) via a retriever \( \mathcal{R} \). The standard RAG constructs the input of LLMs as a concatenation of the documents and the query:
\begin{equation}
    x = [d_1,\cdots,d_N; q],
\end{equation}
and generates \( y \) via LLM:
\begin{equation}
    y = \arg\max_{y'} P(y' \mid x; \theta),
\end{equation}
where \( \theta \) is the parameter of LLM. However, this setup inevitably exposes private documents at inference time, raising serious privacy concerns in many practical scenarios.

In PRAG, given the query $q$ and the retrieved documents $D$, it synthesizes QA pairs from individual document $d_i \in D$ and fine-tunes LLM to obtain its LoRA $\Delta\theta^i$. Then, the resulting LoRAs are summed to inject into LLM for reasoning \( y \) : 
\begin{align}
    y &= \arg\max_{y'} P(y' \mid q; \theta+\Delta\theta), \\
    \Delta\theta &= \sum^N_{i=1} \Delta\theta^i.
\end{align}

\section{Model}
This section introduces the DistilledPRAG model, covering synthetic data construction, knowledge distillation design, and online inference procedures, as shown in Figure~\ref{fig:main}.

\subsection{Synthetic Data Construction}
\label{sec:synthetic}

To supervise the parametric RAG model in answering questions, we construct a large-scale dataset with 289,079 synthetic QA pairs by DeepSeek-V3~\cite{deepseekai2025deepseekv3technicalreport} as the training dataset \( \mathcal{D} \). An example is: “Document: [history of Ada Lovelace] Question: What was Ada Lovelace's main contribution to computer science? Answer: She is credited as the first computer programmer...” The proceeds are as follows:
\begin{enumerate}
    \item We randomly sample 30,000 documents from the 2WikiMultihopQA~\cite{xanh2020_2wikimultihop} training set as $\mathcal{D}_{wiki}$, covering diverse topics and question styles.
    \item For each document \( d_i \in \mathcal{D}_{wiki}\), we use DeepSeek-V3 with carefully designed prompts(see in Appendix~\ref{app:prompt}) to automatically generate 2-5 high-quality question-answer pairs, donated as $\{(d_i, q_{ij}, a_{ij}) \}$ where $2\leq j \leq 5$, totally 139,723 single-document QA pairs into our training dataset \( \mathcal{D} \).
    \item To simulate realistic multi-document retrieval, for each \( d_i \in \mathcal{D}_{wiki}\), we randomly sample a document $d'_i$ from the corpus $\mathcal{D}_{wiki}$ and concatenate them. Then we use the same DeepSeek-V3 with carefully designed prompts (see in Appendix~\ref{app:prompt}) to automatically generate at least 5 additional cross-document QA pairs, focusing on reasoning that spans both documents, donated as $\{(d_i+d'_i, q'_{ij}, a'_{ij}) \}$ where $j \geq 5$, totally 149,356 cross-document QA pairs into \( \mathcal{D} \).
\end{enumerate}
This construction yields a challenging multi-hop QA dataset with broad coverage, which encourages the model to encode document semantics into adapter parameters deeply. For the convenience of a formal definition, we uniformly define the training set as \( \mathcal{D} = \{(D_i, q_i, a_i)\}_{i=1}^{N} \), which includes both single-document and cross-document QA pairs,

\subsection{Knowledge Distillation}
We define the teacher and the student models' inputs and outputs for each training triple $(D_i, q_i, a_i) \in \mathcal{D} $ as:
\begin{itemize}
    \item The \textbf{original input} \( x_i = [D_i; q_i] \) is passed to the standard LLMs \( f_\theta \).
    \item The \textbf{masked input} \( \tilde{x}_i = [\verb+<|doc_mask|>+^{|D_i|}; q_i] \) is passed to the student's LLMs \( f_{\theta + \Delta\theta_i} \). Here, $\verb+<|doc_mask|>+^{|D_i|}$ is formed by repeating the special token \verb+<|doc_mask|>+ to match the length of $|D_i|$, \( \Delta\theta_i = \mathcal{T}_\phi(D_i) \), where $\mathcal{T}_\phi$ is a parameter generator network to map documents $D_i$ to its corresponding LoRA parameters \( \Delta\theta_i\).
\end{itemize}

\subsubsection{Special Token Initialization}
To prevent raw document content from being leaked during inference, we introduce a special token \verb+<|doc_mask|>+ which replaces all tokens in the raw document. Naive initialization (e.g., random values) often leads to unstable training or degraded performance, as they are mismatched with the distribution of the pretrained embedding space( see in Experiments~\ref{sec:special_token}). 
Therefore, we propose a special token initialization that aligns with the first-order and second-order statistics of the model's pretrained vocabulary, thereby promoting stable training. 

Formally, let \( \mathbf{E} \in \mathbb{R}^{V \times h} \) be the embedding matrix of LLMs, where \( V \) is the vocabulary size and \( h \) is the hidden state size. We compute the mean and variance of the embedding distribution as:
\begin{align}
\boldsymbol{\mu} = \frac{1}{V} \sum_{i=1}^{V} \mathbf{E}_i, \quad
\boldsymbol{\sigma} = \sqrt{ \frac{1}{V} \sum_{i=1}^{V} (\mathbf{E}_i - \boldsymbol{\mu})^2 }.
\end{align}
Then, we sample the special token's embedding from this distribution:
\begin{equation}
    \mathbf{e}_{\text{mask}} = \boldsymbol{\mu} + \boldsymbol{\epsilon}, \quad \boldsymbol{\epsilon} \sim \mathcal{N}(\mathbf{0}, \text{diag}(\boldsymbol{\sigma}^2)),
\end{equation}
where $\text{diag}(.)$ is the diagonal matrix.

After initialization, $\mathbf{e}_{\text{mask}}$ is frozen mask and serves as a stable, information-free placeholder used only for document masking, never appearing in queries or answers.

\subsubsection{Parameter Generator}

Here, \( \mathcal{T}_\phi \) denotes a \emph{parameter generator network} that transforms the full documents \( D_i \) into LoRA weights \( \Delta\theta_i \). Specifically, it consists of two components: a document embedding model \( \mathrm{Enc}_\psi \) and a parameter generator \( \mathrm{Gen}_\omega \), such that:
\begin{equation}
    \mathcal{T}_\phi(D_i) = \mathrm{Gen}_\omega\left( \mathrm{Enc}_\psi(D_i) \right),
\end{equation}
where \( \mathrm{Enc}_\psi \) is a pretrained LongT5 encoder~\cite{guo-etal-2022-longt5} to encode documents \( D_i \) into embeddings $\mathbf{E_{D_i}} \in \mathbb{R}^{L \times d}$, where $L$ is the sequence length of $D_i$ and $d$ is the embedding dimension.

Parameter generator $\mathrm{Gen}_\omega$ maps the document embeddings $\mathbf{E_{D_i}}$ to the target LoRA. Specifically, given $\mathbf{E_{D_i}}$ as key and value, a learnable query embeddings $\mathbf{Q} \in \mathbb{R}^{\tilde{L} \times d}$, where $\tilde{L}$ is the number of hidden layers in LLM, we utilize the multi-head cross-attention mechanism to selectively aggregate information from the documents:

\begin{equation}
\mathbf{H}_0 = \text{CrossAttention}(\mathbf{Q},\, \mathbf{E_{D_i}},\, \mathbf{E_{D_i}}).
\end{equation}
Next, the output $\mathbf{H}_0$ undergoes a self-attention-based Transformer encoder, which models dependencies and integrates information within the query sequence itself:
\begin{equation}
\mathbf{H}_1 = \text{SelfAttentionEncoder}(\mathbf{H}_0).
\end{equation}
Finally, a feed-forward network (FFN) maps the encoded representation to the desired LoRA parameter space:
\begin{equation}
\mathcal{T}_\phi(D_i) = \mathrm{FFN}(\mathbf{H}_1).
\end{equation}

This architecture enables efficient extraction, aggregation, and projection of document knowledge into model parameters. Importantly, both the base LLM parameters \( \theta \) and the document encoder parameters \( \psi \) are frozen throughout training, and only the generator $\mathrm{Gen}_\omega$ is updated.

\subsubsection{Training Objectives}
The training objectives of the parameter generator are: generative loss on synthetic QA data, alignment loss of internal representations, and KL divergence of output logits between the student model and the teacher model.

\paragraph{Generative Loss:} Given the documents $D_i$ and the query $q_i$, we minimize the negative log-likelihood for generating the answer \( a_i \) from the masked input $\tilde{x}_i$:
\begin{align}
    \mathcal{L}_{\text{gen}} &= -\log P(a_i \mid \tilde{x}_i;\, \theta + \Delta\theta_i), \\
    \Delta\theta_i &= \mathcal{T}_\phi(D_i).
\end{align}

\paragraph{Internal Alignment Loss:} To align the internal reasoning process, we match their hidden representations at each layer between the student model and the teacher model. Let \( h_{\text{t}}^{(j)} \) and \( h_{\text{s}}^{(j)} \) denote the hidden states from the \(j\)-th hidden layer of teacher model \( f_\theta(x_i) \) and student model \( f_{\theta + \Delta\theta_i}(\tilde{x}_i) \), respectively. The cosine similarity alignment loss for the \(j\)-th layer is defined as:
\begin{equation}
    \mathcal{L}_{\text{cos}}^{(j)} = 1 - \cos(h_{\text{t}}^{(j)},\, h_{\text{s}}^{(j)}).
\end{equation}
The overall alignment loss is then computed as the weighted average of the losses from all layers, where the weight of the \(j\)-th layer is proportional to \(j\):
\begin{equation}
    \mathcal{L}_{\text{cos}} = \frac{\sum_{j=1}^M j\, \mathcal{L}_{\text{cos}}^{(j)}}{\sum_{j=1}^M j}
\end{equation}
where \(M\) is the total number of hidden layers in LLM. We believe that the layers closer to the output are more important.

\paragraph{KL Divergence of Output Logits:} To further ensure output consistency, we align the predictive distributions at the answer tokens \( a_i \) by minimizing the Kullback-Leibler divergence~\cite{KL} between the logits from the student model and the teacher model. Let \( \mathbf{z}_{\text{t}} \) and \( \mathbf{z}_{\text{s}} \) be the pre-softmax logits of the teacher model and the student model for answer positions:
\begin{equation}
    \mathcal{L}_{\text{KL}} = \text{KL} \left( \text{softmax}(\mathbf{z}_{\text{t}}) \;\|\; \text{softmax}(\mathbf{z}_{\text{s}}) \right).
\end{equation}

Finally, we optimize only the generator parameters \( \omega \) via gradient descent over the above objectives: $\mathcal{L}_{\text{gen}}$, $\mathcal{L}_{\text{cos}}$ and $\mathcal{L}_{\text{KL}}$. The base model and document encoder remain frozen during training. The overall objective can be written as:
\begin{equation}
    \min_{\omega} \; \mathbb{E}_{(D_i, q_i, a_i) \sim \mathcal{D}} \left[ \mathcal{L}_{\text{gen}} + \lambda_1 \mathcal{L}_{\text{cos}} + \lambda_2 \mathcal{L}_{\text{KL}} \right],
\end{equation}
where \( \lambda_1, \lambda_2 \) are hyperparameters that balance hidden-states and output-logits alignment. Note that our distillation is not performed between a large teacher model and a smaller student model in the traditional sense, but rather between two different input paradigms.

\subsection{Online Inference}

At inference time, our approach diverges from previous methods such as PRAG and DyPRAG, which require explicit fusion of LoRA parameters from multiple documents. In PRAG, the LoRA A and B matrices generated on each document are concatenated along the LoRA rank dimension. In contrast, DyPRAG directly averages the LoRA A and B weights generated on each document. Instead, we adhere to the same input organization paradigm as in our training phase. During the fusion of LoRA weights, there may be loss or corruption of knowledge and information, as this fusion process is not explicitly modeled during training in their methods. In contrast, our inference paradigm remains consistent with training, helping enhance generalization performance.

Given a user query $q_{\text{test}}$, we first employ a retriever $\mathcal{R}$ to obtain the top-$k$ relevant documents $\{d_1, \ldots, d_k\}$. These documents are concatenated in their original order to form a composite document $d_{\text{inf}} = [d_1; \ldots; d_k]$. To preserve privacy, we mask the entire $d_{\text{inf}}$ using our special token, resulting in the masked input $\tilde{x}_{\text{inf}} = [\texttt{<|doc\_mask|>}^{|d_{\text{inf}}|}; q_{\text{test}}]$. The composite document is then passed to the parameter generator network $\mathcal{T}_\phi$, which computes LoRA adapter weights $\Delta\theta_{\text{inf}} = \mathcal{T}_\phi(d_{\text{inf}})$ conditioned on the composite document. The LoRA-augmented model 
$f_{\theta + \Delta\theta_{\text{inf}}}$
processes the masked input and generates the final answer (prompts in the Appendix~\ref{app:prompt}):
\begin{equation}
    g_{\text{test}} = f_{\theta + \Delta\theta_{\text{inf}}}(\tilde{x}_{\text{inf}}).
\end{equation}

In summary, our online inference procedure is fully aligned with the data format during training and obviates the need for complex parameter fusion schemes commonly required in prior works. During inference, the document is never exposed in plaintext, and only the dynamically generated \( \Delta\theta_i \) is used to condition the model.

\begin{table*}[!t]
  \centering
  \resizebox{\textwidth}{!}{
    \begin{tabular}{llccccccccc}
    \toprule
    \multirow{2}{*}[-0.5ex]{Base LLM} & \multirow{2}{*}[-0.5ex]{Method} & \multicolumn{4}{c}{2WQA} & \multicolumn{2}{c}{HQA} & \multirow{2}{*}[-0.5ex]{PQA} & \multirow{2}{*}[-0.5ex]{CWQ} & \multirow{2}{*}[-0.5ex]{Avg} \\
\cmidrule(lr){3-6}  \cmidrule(lr){7-8}         &       & Compare & Bridge & Inference & Compose & Bridge & Compare &       &       &  \\
    \midrule
    \multirow{4}{*}{LLaMA-1B} & Standard RAG & 34.9  & 32.7  & \textbf{28.2} & \underline{7.9}   & \textbf{18.9} & 27.5  & \underline{17.8}  & 29.1  & 24.6  \\
          & PRAG  & \underline{41.2}  & \underline{41.0}  & 18.9  & 5.5   & 13.9  & \textbf{42.2} & \textbf{21.3} & \underline{31.7}  & \underline{27.0}  \\
          & DyPRAG & 31.3  & 20.9  & 22.3  & 7.2   & 10.4  & 21.3  & 9.7   & 23.2  & 18.3  \\
          & DistilledPRAG & \textbf{42.0} & \textbf{43.1} & \underline{26.3}  & \textbf{14.5} & \underline{14.7}  & \underline{32.6}  & 14.3  & \textbf{38.9} & \textbf{28.3} \\
    \midrule
    \multirow{4}{*}{LLaMA-8B} & Standard RAG & 28.7  & 37.1  & \textbf{31.9} & 8.9   & \textbf{33.6} & \textbf{55.2} & \textbf{33.0} & \underline{41.9}  & \underline{33.8}  \\
          & PRAG  & \textbf{45.1} & 37.0  & 19.8  & 6.9   & 18.6  & 44.7  & 17.6  & 36.2  & 28.2  \\
          & DyPRAG & \underline{44.7}  & \underline{41.9}  & 21.6  & \underline{12.4}  & 14.9  & 47.1  & 12.4  & 41.4  & 29.6  \\
          & DistilledPRAG & 41.3  & \textbf{45.6} & \underline{30.1}  & \textbf{16.2} & \underline{26.9}  & \underline{54.2}  & \underline{25.6}  & \textbf{49.0} & \textbf{36.1} \\
    \midrule
    \multirow{3}{*}{Mistral-7B} & Standard RAG & 26.7 &	\underline{28.1} &	\underline{26.7} &	\underline{6.8} &	\underline{16.2} &	23.6 &	\textbf{18.2} &	18.4 &	20.6 \\
        & PISCO & \underline{28.9} &	27.9 &	\textbf{27.9} &	6.6 & \textbf{16.8} &	\underline{24.0} &	\underline{14.6} &	\textbf{26.0} &	\underline{21.6} \\
        & DistilledPRAG  & \textbf{34.0}&\textbf{35.9} &25.8&\textbf{10.2}&15.4&\textbf{26.4}&12.6&\underline{24.7}&\textbf{23.1} \\
                                
    \bottomrule
    \end{tabular}%
    }
    \caption{\label{tab:main} Overall F1(\%) performance of DistilledPRAG and baselines on 2WQA, HQA, PQA and CWQ datasets. Bold indicates the best performance, and underlined indicates the second best. Our performance on Qwen-14B can be seen in Appendix~\ref{app:appLarger}}
\end{table*}%

\section{Experiments}

To demonstrate the performance of DistilledPRAG, we define the in-domain benchmark and OOD datasets across four datasets and conduct comparisons with baselines.

\subsection{Experimental Setup}

\subsubsection{Datasets and Metrics}
We evaluate question-answering performance using the F1 score(\%) for our DistilledPRAG method and baselines on four open-domain QA datasets: 2WikiMultihopQA(2WQA)~\cite{xanh2020_2wikimultihop}, HotpotQA(HQA)~\cite{yang-etal-2018-hotpotqa}, PopQA(PQA)~\cite{mallen-etal-2023-trust-popqa}, and ComplexWebQuestions(CWQ)~\cite{talmor-berant-2018-Complexwebqa}. For a fair comparison between DistilledPRAG and baselines, following PRAG and DyPRAG, we use the same first 300 questions from the dev split of each sub-task dataset as the test set. \textbf{But their training sets are different.} PRAG requires no training and only loads offline-generated LoRA parameters at test time. DyPRAG proposes to train its parameter translator on questions 301-600 from each sub-task's dev split, which has a distribution similar to the test set. Notably, our DistilledPRAG only use the training split of 2WQA as a training dataset, and test on the 2WQA dev set as the in-domain benchmark, while HQA, PQA, and CWQ dev sets serve as OOD datasets for generalization evaluation. Therefore, \textbf{the generalization evaluation of our model is more rigorous}, demonstrating the effectiveness of our method. Details in Appendix~\ref{app:expdm}.

\subsubsection{Implementation Details}
We conduct experiments using two LLM backbones: \textbf{Llama-3.2-1B-Instruct} and \textbf{LLaMA3-8B-Instruct}, and compare with PISCO using \textbf{Mistral-7B-Instruct-v0.2}. All experiments run in PyTorch on NVIDIA A100 80GB PCIE and RTX PRO 6000 Blackwell GPUs. The parameter generator uses the \textbf{LongT5 encoder} as its embedding model. Training uses AdamW with a learning rate of $10^{-4}$, 10\% warm-up, a polynomial scheduler ending at $10^{-6}$, batch size 4, one epoch, LoRA rank 2, and LoRA alpha 32. We set $\lambda_1 = 0.5$ and $\lambda_2 = 0.1$, keeping all other hyperparameters at defaults. In inference, we disable sampling \verb+do_sample=False+ and use greedy decoding for deterministic outputs.

\subsubsection{Baselines}
For DistilledPRAG and all baselines, we adopt a standard BM25 retriever~\cite{bm25} to fetch the top-3 documents for each question. All baselines are \textbf{Standard RAG}, \textbf{PRAG}~\cite{su2025parametricretrievalaugmentedgeneration}, \textbf{DyPRAG}~\cite{tan2025dynamicparametricretrievalaugmented}, \textbf{PISCO}~\cite{louis2025piscoprettysimplecompression}. See Appendix~\ref{app:baseline} for baseline details and Appendix~\ref{app:appRetriver} for retriever and top-k analysis.

\subsection{Main Results}
\label{sec:main_results}
\begin{table*}[!t]
  \centering
  \resizebox{\textwidth}{!}{
    \begin{tabular}{lccccccccc}
    \toprule
    \multirow{2}[4]{*}[0.5ex]{Method} & \multicolumn{4}{c}{2WQA}      & \multicolumn{2}{c}{HQA } & \multirow{2}[4]{*}{PQA} & \multirow{2}[4]{*}{CWQ} & \multirow{2}[4]{*}{Avg} \\
\cmidrule(lr){2-5}  \cmidrule(lr){6-7}         & Compare & Bridge & Inference & Compose  & Bridge & Compare &       &       &  \\
    
    \midrule
    DistilledPRAG  & \textbf{41.3} & 45.6  & 30.1  & 16.2  & \textbf{26.9} & \textbf{54.2} & \textbf{25.6} & \textbf{49.0} & \textbf{36.1} \\
    
    w/o $\mathcal{L}_{\text{cos}}$ &     41.0  & \textbf{46.3} & 29.2  & 15.7  & 25.8  & 51.2  & 22.8  & 46.0  & 34.7  \\
    
    w/o $\mathcal{L}_{\text{KL}}$ &     33.1  & 33.9  & 29.5  & 16.5  & 24.5  & 42.5  & 24.5  & 44.6  & 31.1  \\

    w/o $\mathcal{L}_{\text{cos}}$, $\mathcal{L}_{\text{KL}}$ &     30.5  & 30.5  & \textbf{30.2} & \textbf{17.2} & 24.3  & 39.8  & 23.3  & 44.5  & 30.0  \\

    \bottomrule
    \end{tabular}%
    }
    \caption{\label{tab:ablation-loss} Ablation study of alignment losses based on LLaMA3-8B-Instruct backbone model.}
\end{table*}%

\begin{table*}[!t]
  \centering
  \resizebox{\textwidth}{!}{
    \begin{tabular}{lccccccccc}
    \toprule
    \multirow{2}[4]{*}{\vspace{1.0ex}Method} & \multicolumn{4}{c}{2WQA}      & \multicolumn{2}{c}{HQA } & \multirow{2}[4]{*}{PQA} & \multirow{2}[4]{*}{CWQ} & \multirow{2}[4]{*}{Avg} \\
\cmidrule(lr){2-5}  \cmidrule(lr){6-7}         & Compare & Bridge & Inference & Compose  & Bridge & Compare &       &       &  \\
    \midrule

    DistilledPRAG  & \textbf{42.0} & 43.1 & 26.3  & \textbf{14.5} & \textbf{14.7}  & \textbf{32.6} & 14.3  & \textbf{38.9} & \textbf{28.3} \\
    Llama-Synthesis & 38.2  & \textbf{44.6}  & 24.3  & 14.1  & \textbf{14.7} & 29.4  & \textbf{18.6} & 36.9  & 27.6  \\
    Single-Document & 33.8  & 42.4  & \textbf{27.0} & 12.9  & \textbf{14.7}  & 29.7 &	12.7 &	38.2 &	26.4 \\

    \bottomrule
    \end{tabular}%
    }
    \caption{\label{tab:ablation-synthesis} The impact of QA synthesis on model performance using Llama-3.2-1B-Instruct. Llama-Synthesis uses Llama-3.2-1B-Instruct (vs. DeepSeek-V3) to generate QA data. Single-Document uses only single-document QA pairs of training dataset $\mathcal{D}$(see in Section~\ref{sec:synthetic}).}
\end{table*}%

Table~\ref{tab:main} reports QA performance for DistilledPRAG and several baselines across datasets and backbone LLMs. DistilledPRAG attains the best or second-best results on most sub-tasks and consistently yields the highest average F1 score. With the LLaMA-8B backbone, it achieves the top average F1, surpassing standard RAG, PRAG, and DyPRAG by 2.3, 7.9, and 6.5, respectively. 

Notably, DistilledPRAG is trained only on the 2WQA training set, whereas the baseline DyPRAG is trained on a similar distribution of data with test data~\footnote{DyPRAG proposes to train its parameter translator on questions 301-600 from each sub-task's dev split, which has a distribution similar to the test set.}.  Nevertheless, DistilledPRAG shows strong OOD generalization on HQA, PQA, and CWQ, achieving a leading 49.0 on CWQ. Similar patterns hold for LLaMA-1B and LLaMA-8B, where DistilledPRAG consistently matches or surpasses the strongest baselines in both in-domain and OOD settings. Overall, DistilledPRAG demonstrates robust transferability and practical effectiveness for real-world open-domain QA.

\subsection{Analysis}

\subsubsection{Alignment Loss}
Table~\ref{tab:ablation-loss} shows that removing either alignment loss $\mathcal{L}_{\text{cos}}$ or $\mathcal{L}_{\text{KL}}$ reduces performance. Dropping $\mathcal{L}_{\text{cos}}$ decreases accuracy by 3.9\%, while removing $\mathcal{L}_{\text{KL}}$  leads to a larger 13.9\% drop. Eliminating both results in a 16.9\% decline. These findings indicate that aligning hidden states and logits is crucial for capturing the teacher model’s representations, improving semantic sensitivity, and enhancing generalization to unseen or out-of-distribution data.


\subsubsection{QA Synthesis}

\begin{table*}[!t]
  \centering
  \resizebox{0.9\textwidth}{!}{
    \begin{tabular}{lccccccccc}
    \toprule
    \multirow{2}[4]{*}[0.5ex]{Method} & \multicolumn{4}{c}{2WQA}      & \multicolumn{2}{c}{HQA } & \multirow{2}[4]{*}{PQA} & \multirow{2}[4]{*}{CWQ} & \multirow{2}[4]{*}{Avg} \\
\cmidrule(lr){2-5}  \cmidrule(lr){6-7}         & Compare & Bridge & Inference & Compose  & Bridge & Compare &       &       &  \\
    \midrule

    DistilledPRAG  & \textbf{42.0} & \textbf{43.1} & \textbf{26.3} & \textbf{14.5} & \textbf{14.7} & 32.6  & 14.3  & \textbf{38.9} & \textbf{28.3} \\
    No-token & 35.5  & 29.8  & 25.3  & 12.5  & 13.6  & 23.3  & 13.3  & 33.5  & 23.3  \\
    Random-token & 39.2  & 41.6  & 22.6  & 9.0   & 9.9   & \textbf{39.3} & 9.2   & 35.5  & 25.8  \\
    Trainable-token & 31.9  & 39.2  & 22.8  & 8.4   & 13.7  & 25.3  & \textbf{15.3} & 31.7  & 23.5  \\
    
    \bottomrule
    \end{tabular}%
    }
    \caption{\label{tab:ablation-token} The impact of special tokens on masked documents using Llama-3.2-1B-Instruct: No-token deletes content without masking; Random-token uses default Transformer initialization for special token; Trainable-token has a learnable special token embedding.
    }
\end{table*}%

Table~\ref{tab:ablation-synthesis} shows the impact of DeepSeek synthetic data and cross-document augmentation. Using QA pairs generated by Llama-3.2-1B leads to a 0.7 average performance drop, suggesting that synthetic QA quality is essential for training the parameter generator. High-quality QA pairs capture key facts and reasoning patterns, providing stronger supervision (see Appendix~\ref{app:judge-psg}). Superficial or incomplete QA generation produces LoRA parameters that miss key document knowledge, weakening downstream performance.
Moreover, when training only on single-document QA pairs, performance drops by 1.9 because the training data lacks knowledge diversity. Cross-document QA synthesis adds questions that require reasoning across multiple sources, helping the model learn broader, more transferable representations. Without this augmentation, the model overfits to narrow, document-specific patterns and struggles with complex or open-domain queries. Additionally, we also compare DistilledPRAG with baselines on the same synthesizer and data size in Appendix~\ref{app:appEqualQA}.
%



\subsubsection{Special Token}
\label{sec:special_token}
Table~\ref{tab:ablation-token} compares several mask-token configurations: no token, a randomly initialized token, and a trainable token.
(1) Removing mask tokens causes a large F1 drop (17.7\%), showing that preserving the model’s structural paradigm is crucial for effective parameter-generator training.
(2) Using a randomly initialized fixed token $\mathbf{e}_{\text{mask}} \sim \mathcal{N}\left(0, \sigma^2 I\right), \sigma=0.02$ yields an 8.9\% lower F1 than ours. Because such a token lacks semantic grounding, the model struggles to interpret it. In contrast, our token is sampled using the mean and variance of all vocabulary embeddings, aligning it with the pretrained embedding space and stabilizing training.
(3) Making the mask token trainable results in a notable performance drop, likely because continual changes to the token destabilize document representations.\footnote{Document-specific mask tokens are impractical here, as each document has fewer than 15 QA pairs, insufficient for effective token training.}


\subsubsection{Inference Latency}

\begin{table}[!t]
  \centering
  
  \resizebox{\columnwidth}{!}{
    \begin{tabular}{lccc}
    \toprule
    \multirow{2}[4]{*}[0.5ex]{Method} & \multicolumn{3}{c}{Latency} \\
\cmidrule(lr){2-4}  & LLaMA-1B & LLaMA-8B & Mistral-7B\\
    \midrule
    Standard RAG & 0.14 & 0.52 & 0.76\\
    PRAG  & 0.6(+18)   & 1.18(+100) & -\\
    DyPRAG & 0.52  & 0.8 &-\\
    PISCO & - & - & 0.92 \\
    DistilledPRAG  & 0.3   & 0.81 & 1 \\
    \bottomrule
    \end{tabular}%
    }
    \caption{\label{tab:latency} Average inference latency (s). For PRAG, extra time (in brackets) is added for offline LoRA synthesis and training when new documents are introduced.
 }
\end{table}%
Table~\ref{tab:latency} shows the average inference latency of all methods under identical hardware (RTX PRO 6000 and Intel Xeon Gold 5318Y). Our method achieves the lowest latency for small models, except for RAG, because it requires only one round of parameter generation. In contrast, DyPRAG needs three rounds of parameter generation and aggregation, and PRAG repeatedly loads offline document parameters, which significantly increases overhead. Loading LoRA parameters accounts for 33.9\%, 38.9\%, and 6.9\% of total inference time for PRAG, DyPRAG, and DistilledPRAG, respectively. As model size grows, our latency increases due to special tokens matching the length of the original documents. Still, our inference process is the same as standard RAG, and our goal is to match RAG-level performance without plaintext documents rather than to optimize speed.


\subsubsection{Overlap Between Train and Test}
\begin{table}[!t]
  \centering
  \resizebox{\columnwidth}{!}{
    \begin{tabular}{ccccc}
    \toprule
          & \multicolumn{1}{c}{2WQA} & \multicolumn{1}{c}{HQA} & \multicolumn{1}{c}{PQA} & \multicolumn{1}{c}{CWQ} \\
    \midrule
    DyPRAG & 19.9  & 17.1  & 17.5  & 25.4  \\
    Ours  & 19.8  & 17.5  & 17.8  & 16.6  \\
    \bottomrule
    \end{tabular}%
  }
\caption{\label{tab:leakage} The overlap (\%) between documents used in training and those retrieved from the test set.}
\end{table}%

To further validate the generalization of DistilledPRAG and eliminate the risk of test set leakage, we estimated the maximum Jaccard similarity~\cite{jaccard} between each test document and all training documents using an efficient MinHash and Locality‑Sensitive Hashing(LSH) approach. For each test document, we recorded the highest similarity with any training document and averaged these scores. As shown in Table~\ref{tab:leakage}, the average maximum similarity across four datasets remains below 20\%. This demonstrates limited textual overlap and confirms the robustness and authenticity of our method’s generalization. Additionally, we add analysis of reconstruction attack in Appendix~\ref{app:appAttack}.

\section{Conclusion}
In this work, we present DistilledPRAG, a novel parametric RAG model that addresses the generalization and efficiency limitations of existing approaches such as PRAG and DyPRAG. By aligning both input structure and parameter activations with a standard RAG teacher model, DistilledPRAG leverages knowledge distillation to achieve robust RAG-style reasoning without access to plaintext documents. Experimental results on four public QA datasets demonstrate that DistilledPRAG significantly outperforms strong baselines in OOD settings, even when trained on a single dataset. In future work, we plan to extend DistilledPRAG to support multi-modal inputs and explore its applicability in settings with more complex reasoning and open-domain generation tasks.

\section*{Limitations}

Despite its promising performance, DistilledPRAG remains an approximation of standard RAG. While our internal alignment strategy improves reasoning consistency, the parametric nature of LoRA representations may still fall short in capturing nuanced cross-document interactions, especially in complex multi-hop scenarios. Furthermore, the privacy-preserving design by avoiding transmission of plaintext documents inevitably introduces a privacy-utility tradeoff. Encoding documents into LoRA may lose fine-grained contextual details, limiting answer accuracy on knowledge-intensive queries. Future work may explore richer alignment objectives and more expressive methods to further close the gap between parametric and standard RAG performance.



\bibliography{custom}

\appendix
\section{Experimental Settings}
\subsection{Datasets and Metrics} \label{app:expdm}
We evaluate question-answering performance using the F1 score(\%) for our DistilledPRAG method and baselines on four open-domain QA datasets: 2WikiMultihopQA(2WQA)~\cite{xanh2020_2wikimultihop}, HotpotQA(HQA)~\cite{yang-etal-2018-hotpotqa}, PopQA(PQA)~\cite{mallen-etal-2023-trust-popqa}, and ComplexWebQuestions(CWQ)~\cite{talmor-berant-2018-Complexwebqa}. 2WQA and HQA evaluate multi-hop reasoning by requiring integration of information from multiple sources. PQA tests factual recall and entity disambiguation. CWQ assesses multi-step reasoning over web-based content. The 2WQA and HQA datasets categorize questions by reasoning type, with four sub-tasks for 2WQA and two for HQA.

For a fair comparison between DistilledPRAG and baselines, following PRAG and DyPRAG, we use the same first 300 questions from the dev split of each sub-task dataset as the test set. \textbf{But their training sets are different.} PRAG requires no training and only loads offline-generated LoRA parameters at test time. DyPRAG proposes to train its parameter translator on questions 301-600 from each sub-task's dev split, which has a distribution similar to the test set. Notably, our Distilled PRAG only use the training split of 2WQA as a training dataset, and test on the 2WQA dev set as the in-domain benchmark, while HQA, PQA, and CWQ dev sets serve as OOD datasets for generalization evaluation. Therefore, \textbf{the generalization evaluation of our model is more rigorous}, demonstrating the effectiveness of our method.


\subsection{Baselines} \label{app:baseline}
For DistilledPRAG and all baselines, we adopt a standard BM25 retriever~\cite{bm25} to fetch the top-3 documents for each question. Our method and all baselines are as follows:
\begin{itemize}
    \item \textbf{Standard RAG}: Using the retrieved documents as in-context input for backbone LLM.
    \item \textbf{PRAG}~\cite{su2025parametricretrievalaugmentedgeneration}: Parameterizing each document from the corpus into LoRA weights by offline synthesizing QA pairs and fine-tuning LLM. During inference, they retrieve the corresponding LoRAs for each query-relevant document and sum them according to the LoRA rank dimension.  
    \item \textbf{DyPRAG}~\cite{tan2025dynamicparametricretrievalaugmented}: During inference, dynamically generating document-specific LoRA using a translator network for each document, and averaging them as the aggregated document representation.
    \item  \textbf{PISCO}~\cite{louis2025piscoprettysimplecompression}: Optimized from COCOM~\cite{cocom}, compressing each retrieved document into a few embeddings and concatenating them with the input's embedding for backbone LLM.
    \item \textbf{DistilledPRAG}: The parameter generator directly encodes the retrieved top-3 documents to the unified multi-document LoRA, without additional LoRA aggregation.
\end{itemize}

\subsection{Prompts Used in Our Experiments}
\label{app:prompt}
\begin{figure}[h]
\centering
\includegraphics[width=1\columnwidth]{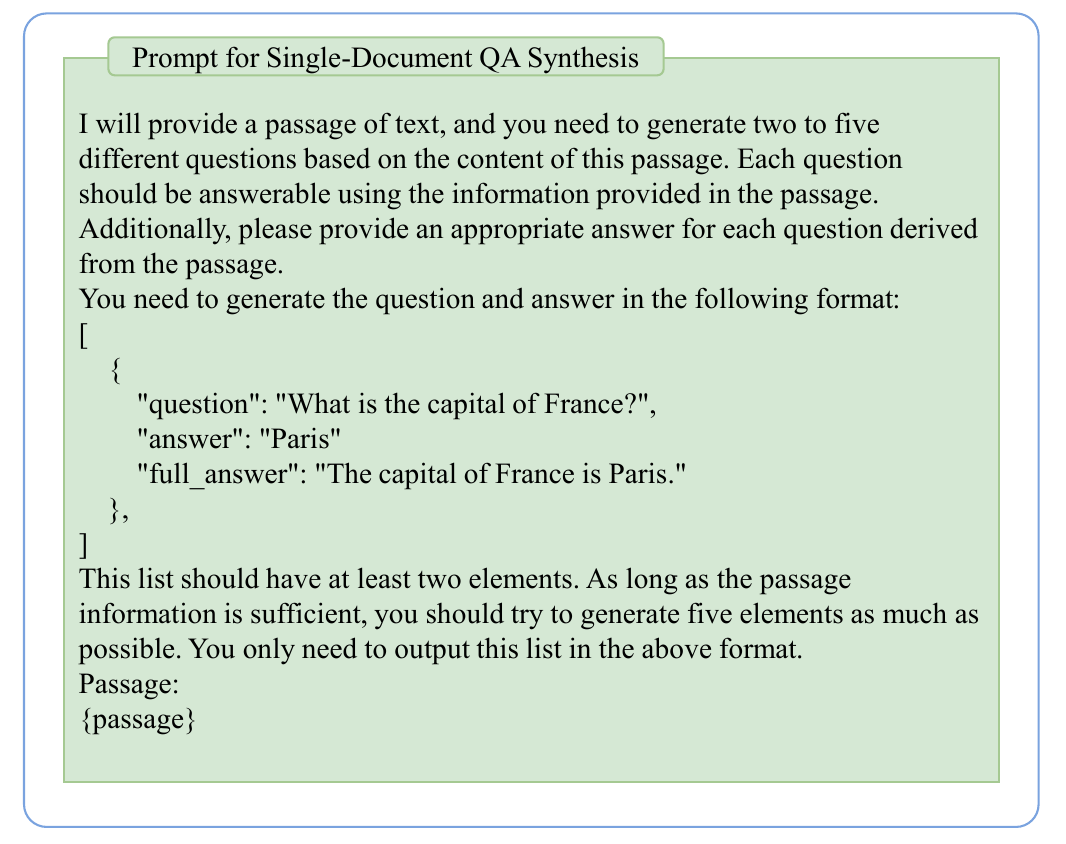}
\caption{\label{fig:prompt1} Prompt for single-document QA synthesis.}
\end{figure}

\begin{figure}[h]
\centering
\includegraphics[width=1\columnwidth]{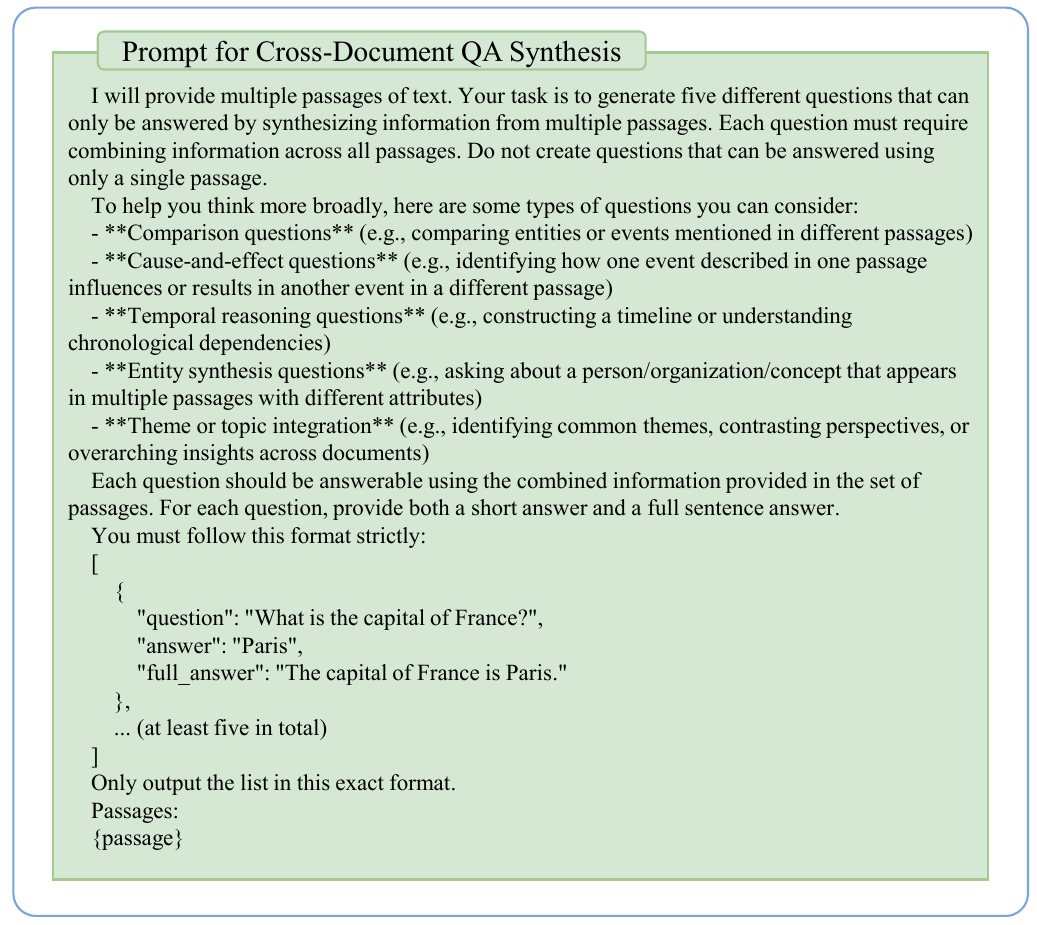}
\caption{\label{fig:prompt2} Prompt for single-document QA synthesis.}
\end{figure}

\begin{figure}[h]
\centering
\includegraphics[width=1\columnwidth]{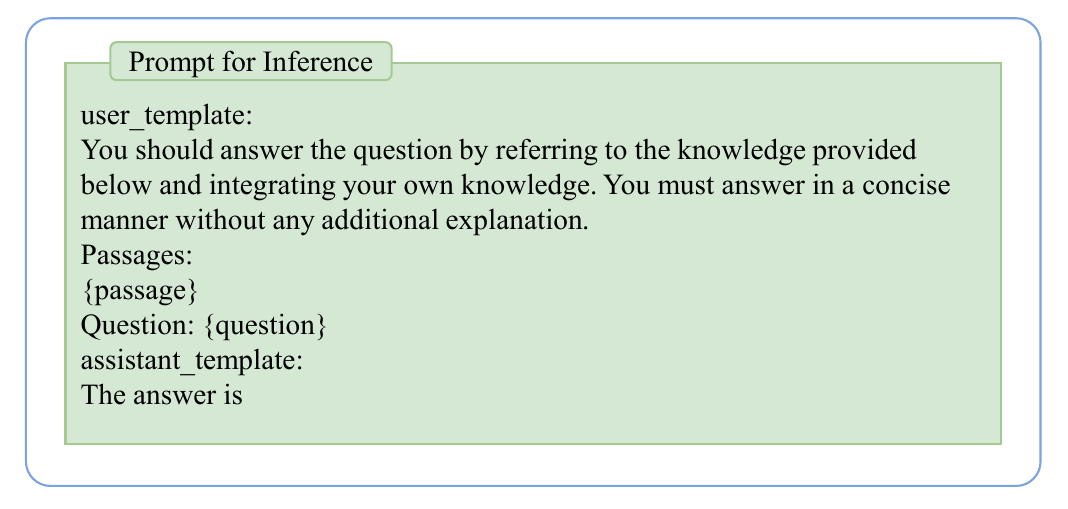}
\caption{\label{fig:prompt3} Prompt for inference.}
\end{figure}

Figures~\ref{fig:prompt1}, \ref{fig:prompt2}, and \ref{fig:prompt3} show the prompts we use for single-document QA synthesis, cross-document QA synthesis, and inference.

\section{Additional Analysis}

\subsection{Cross-document Testing}
\label{app:cross-psg}

We extract 100 challenging cross-document QA pairs synthesized by DeepSeek-V3 as test data to evaluate the robustness of different methods under strong cross-document reasoning requirements. As shown in Table~\ref{tab:cross-psg}, DistilledPRAG achieves an F1 score of 30.9, which is remarkably close to the performance of standard RAG (34.6). This indicates that DistilledPRAG preserves most of RAG’s cross-document reasoning capability even without accessing the original documents at inference time.

In contrast, PRAG and DyPRAG show severe degradation in this setting, with F1 scores dropping to 11.8 and 5.0, respectively. Such near-collapse suggests that their LoRA aggregation strategies struggle to capture the multi-hop relationships required for cross-document questions. The sharp performance gap further demonstrates that the distillation procedure in DistilledPRAG, particularly the cross-document QA design, plays a critical role in maintaining reasoning ability across dispersed evidence.

\begin{table}[!t]
  \centering
  \resizebox{\columnwidth}{!}{
    \begin{tabular}{ccccc}
    \toprule
          & \multicolumn{1}{c}{RAG} & \multicolumn{1}{c}{PRAG} & \multicolumn{1}{c}{DyPRAG} & \multicolumn{1}{c}{DistilledPRAG} \\
    \midrule
    F1 & 34.6  & 11.8  & 5.0  & 30.9  \\
    \bottomrule
    \end{tabular}%
  }
\caption{\label{tab:cross-psg}Performance on LLaMA3-8B-Instruct in strong cross-document scenarios.}
\end{table}%

\subsection{Retrieval Quality}
\label{app:appRetriver}
\begin{figure}[!t]
\centering
\includegraphics[width=\columnwidth]{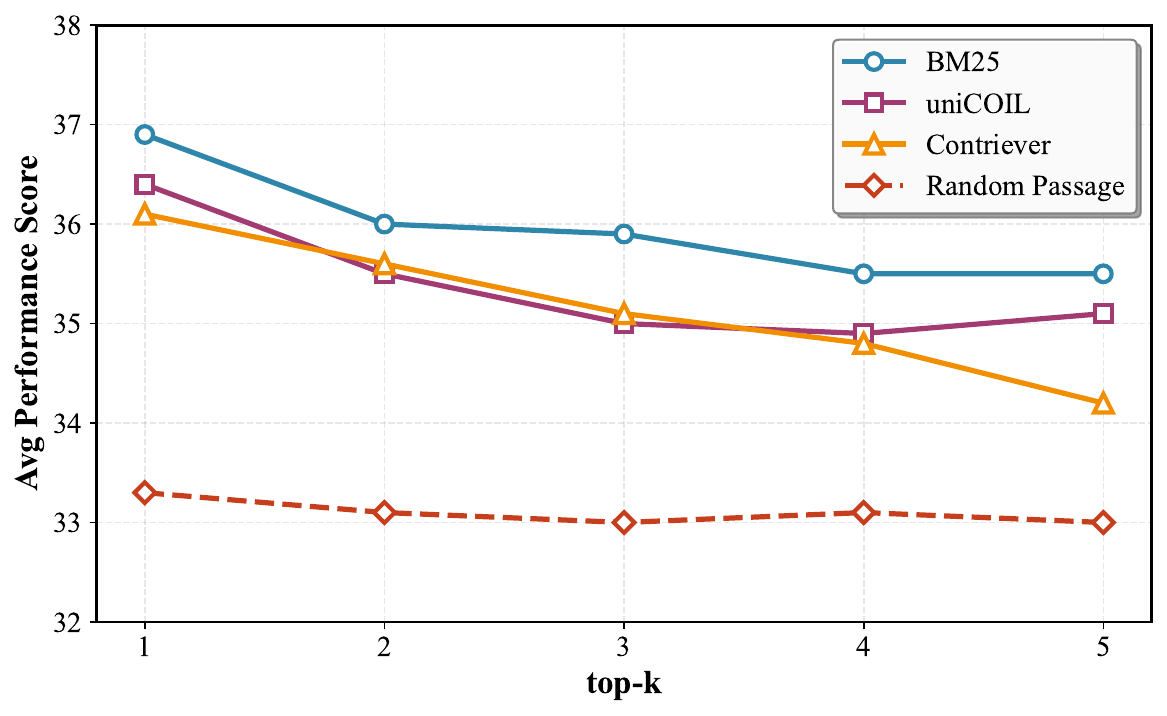}
\caption{\label{fig:retrieval} Performance under different search engines and different top-k values.}
\end{figure}
From Figure~\ref{fig:retrieval}, across BM25, uniCOIL~\cite{unicoil}, and Contriever~\cite{contriever}, we observe a consistent pattern: performance decreases as top-k increases from 1 to 5, and drops further when passages are replaced with irrelevant documents. This confirms that retrieval relevance plays a central role in DistilledPRAG.

The decline with larger k likely stems from DistilledPRAG’s design: all retrieved passages are merged into a single input before producing one LoRA update. Increasing k, therefore, introduces longer and noisier merged contexts, causing weakly relevant or irrelevant content to be directly encoded into parameters. Compared with standard RAG, where distractors mainly dilute attention, this parameterization is inherently more sensitive to heterogeneous document sets. Handling very long or noisy document clusters is a promising direction for future exploration. BM25 performs best because its lexical retrieval introduces fewer misleading distractors, whereas dense retrievers more often retrieve semantically similar but non-answer-bearing passages that can "pollute" the update. The irrelevant-document results further support this explanation: performance decreases but remains above collapse, indicating that DistilledPRAG retains some general reasoning ability while still reflecting the quality of retrieved evidence.

\subsection{Larger Model}
\label{app:appLarger}
\begin{table}[!t]
  \centering
  \resizebox{\columnwidth}{!}{
    \begin{tabular}{lccccc}
    \toprule
    \multirow{2}[4]{*}{\vspace{1.0ex}Method} & \multicolumn{4}{c}{2WQA}  & \multirow{2}[4]{*}{Avg} \\
\cmidrule(lr){2-5}     & Compare & Bridge & Inference & Compose  & \\
    \midrule
    PRAG  & 38.0 & 38.7 & 32.1  & 9.7 & 29.6 \\
    DistilledPRAG & 45.7  & 45.8  & 25.1 & 12.5 & 32.3 \\

    \midrule
    \multirow{2}[4]{*}{\vspace{1.0ex}Method} & \multicolumn{2}{c}{HQA } & \multirow{2}[4]{*}{PQA} & \multirow{2}[4]{*}{CWQ} & \multirow{2}[4]{*}{Avg} \\
  \cmidrule(lr){2-3}   & Bridge & Compare &       &   &  \\
    \midrule

    PRAG  & 35.0  & 52.1 & 38.2  & 41.3 & 41.7 \\
    DistilledPRAG & 22.1  & 60.2 & 25.4 &	48.6 &	39.1 \\
    \bottomrule
    \end{tabular}%
    }
    
    \caption{\label{tab:qwen3}Performance on Qwen-14B.}
\end{table}%
To explore the potential of DistilledPRAG on newer, larger, and architecturally different models, we further conduct experiments on Qwen3-14B. This evaluation allows us to examine whether the benefits observed on LLaMA3-8B generalize to a stronger base model. As shown in Table~\ref{tab:qwen3}, DistilledPRAG achieves an average performance that is nearly on par with standard RAG across all benchmarks. These results suggest that DistilledPRAG retains much of its advantage even when scaled to larger models, and that its design is broadly compatible with architectures beyond LLaMA-family models.

\subsection{Quality Evaluation of Synthetic QA}
\label{app:judge-psg}
\begin{figure}[!t]
    \centering
    \begin{subfigure}{\columnwidth}
        \centering
        \includegraphics[width=\columnwidth]{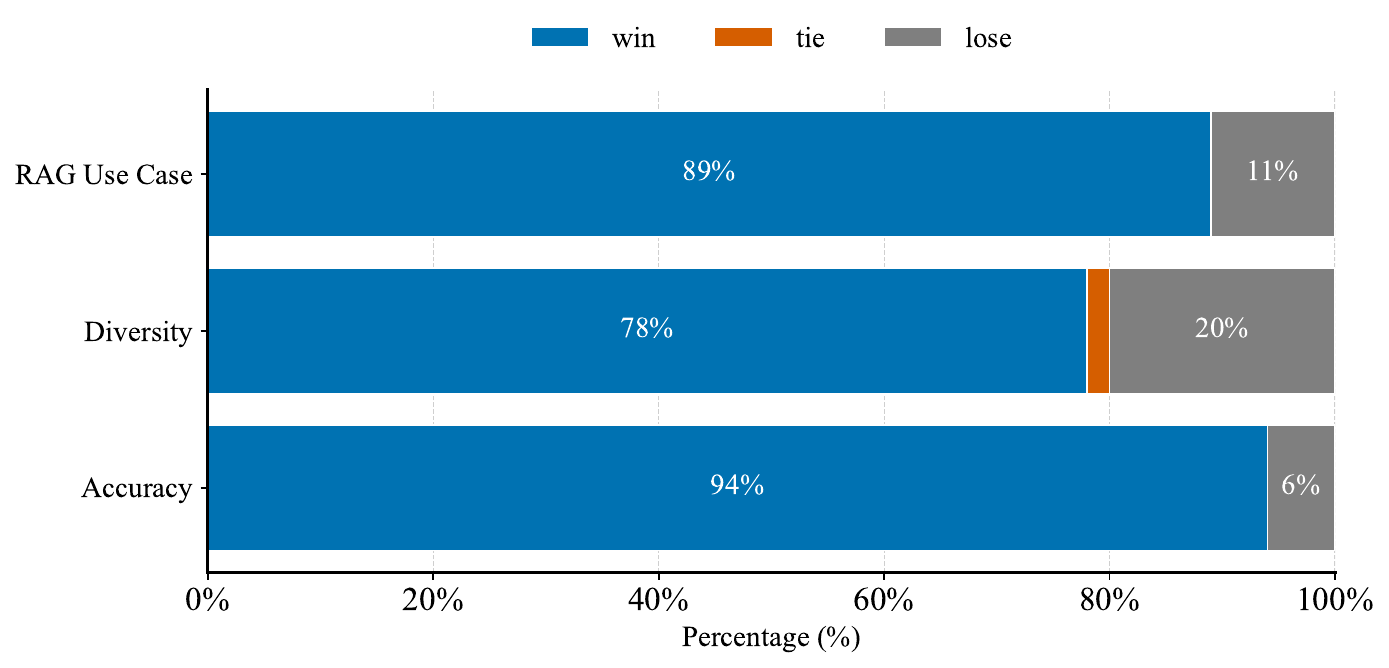}
        \caption{}
        \label{fig:judge-psg-a}
    \end{subfigure}

    \begin{subfigure}{\columnwidth}
        \centering
        \includegraphics[width=\columnwidth]{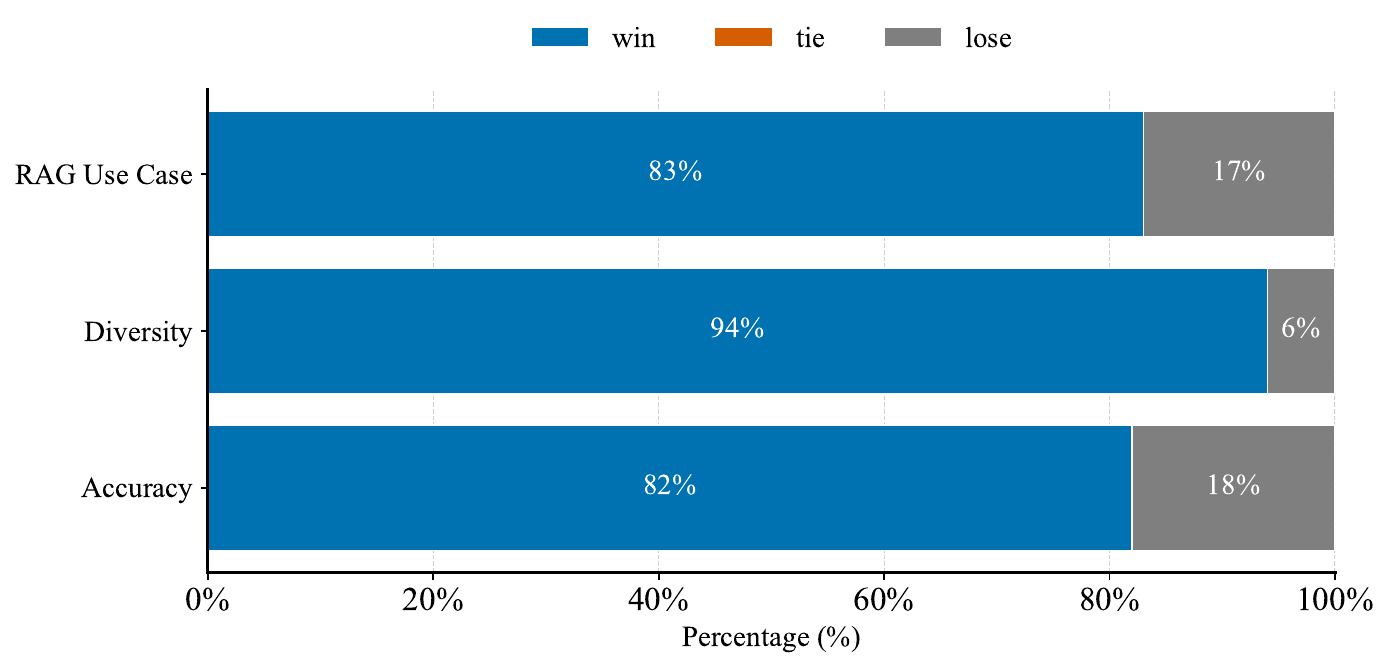}
        \caption{}
        \label{fig:judge-psg-b}
    \end{subfigure}
    
    \caption{Comparison between DeepSeek-V3 and Llama3-8B on synthetic QA quality across three dimensions: accuracy, diversity, and fitness for RAG use cases. (a) Results in the single-document setting. (b) Results in the cross-document setting.}
    \label{fig:judge-psg}
\end{figure}

In this section, we compare the quality of synthetic QA pairs generated by DeepSeek-V3 and Llama3-8B. We employ GPT-5 as the evaluator and assess the QA pairs along three dimensions: accuracy, diversity, and fitness for RAG use cases. As shown in Figure~\ref{fig:judge-psg}, DeepSeek-V3 consistently produces substantially higher-quality QA pairs than Llama3-8B in both single-document and cross-document scenarios.

Specifically, DeepSeek-V3 wins 89\% / 78\% / 94\% of comparisons in the single-document setting and 83\% / 94\% / 82\% in the cross-document setting. These consistently high margins indicate that DeepSeek-V3 not only generates more accurate and varied questions but also produces QA pairs that better reflect real RAG retrieval. The strong performance in cross-document assessments is particularly notable, as it suggests that DeepSeek-V3 excels at producing compositional and multi-hop reasoning questions that are crucial for robust RAG training.

\subsection{Same Synthesizer and Data size}
\label{app:appEqualQA}

\begin{table*}[!t]
  \centering
  \resizebox{\textwidth}{!}{
    \begin{tabular}{lccccccccc}
    \toprule
    \multirow{2}[4]{*}{\vspace{1.0ex}Method} & \multicolumn{4}{c}{2WQA}      & \multicolumn{2}{c}{HQA } & \multirow{2}[4]{*}{PQA} & \multirow{2}[4]{*}{CWQ} & \multirow{2}[4]{*}{Avg} \\
\cmidrule(lr){2-5}  \cmidrule(lr){6-7}         & Compare & Bridge & Inference & Compose  & Bridge & Compare &       &       &  \\
    \midrule

    PRAG  & 30.9 & 31.4 & 19.3  & 7.3 & 16.2  & 33.8 & 15.7  & 33.8 & 23.6 \\
    PRAG(ds) & \textbf{46.8} & \textbf{46.0} & 20.5  & 10.1 & 18.3  & 30.8 & 19.3  & 36.4 & 28.5 \\
    DyPRAG & 24.9  & 18.5  & 20.6  & 7.6  & 15.5 & 34.5  & 19.5 & 39.2  & 22.5  \\
    DyPRAG(ds) & 33.0  & 19.5  & 20.1 & 9.1  & 16.6  & 42.2 &	10.4 &	35.3 &	23.3 \\
    DistilledPRAG & 32.2  & 41.1  & \textbf{29.6} & \textbf{17.5}  & \textbf{26.3}  & \textbf{47.8} &	\textbf{24.3} &	\textbf{46.1} &	\textbf{33.1} \\

    \bottomrule
    \end{tabular}%
    }
    \caption{\label{tab:fair-base}Fairer Baseline Comparison on LLaMA3-8B-Instruct. The "ds" setting refers to using DeepSeek-V3 for QA synthesis.}
\end{table*}%

Although the previous experiments already demonstrate the effectiveness of our method, we further conduct a QA pair-controlled comparison, ensuring that all methods use the same number of QA pairs per document. We adjust the training documents to exactly match those used in DyPRAG, and we ensure that PRAG, DyPRAG, and DistilledPRAG use the same number of synthetic QA pairs per document (10 each). In our case, half of the QA pairs are used for single-document training and the other half for cross-document training. As shown in Table~\ref{tab:fair-base}, DistilledPRAG still achieves the best performance, further indicating that the cross-document QA design and the introduction of mask tokens play a crucial role.

As shown in Table~\ref{tab:fair-base}, DistilledPRAG continues to outperform all baselines across nearly all categories, even when the training supervision is strictly matched. PRAG and DyPRAG exhibit modest improvements when using DeepSeek-V3-generated QA pairs (“ds”), but their gains remain limited, suggesting that their LoRA aggregation and parameterization strategies do not fully exploit cross-document reasoning signals. In contrast, DistilledPRAG shows substantial and consistent improvements in both 2WQA and HQA subsets, indicating that its cross-document QA design and the use of mask tokens enable the model to better capture relational and compositional reasoning patterns. This controlled comparison further confirms that the benefits of DistilledPRAG stem from its learning mechanism rather than from differences in QA quality and quantity.

\subsection{Reconstruction Attack}
\label{app:appAttack}
\begin{figure}[!t]
    \centering
    \begin{subfigure}{\columnwidth}
        \centering
        \includegraphics[width=\columnwidth]{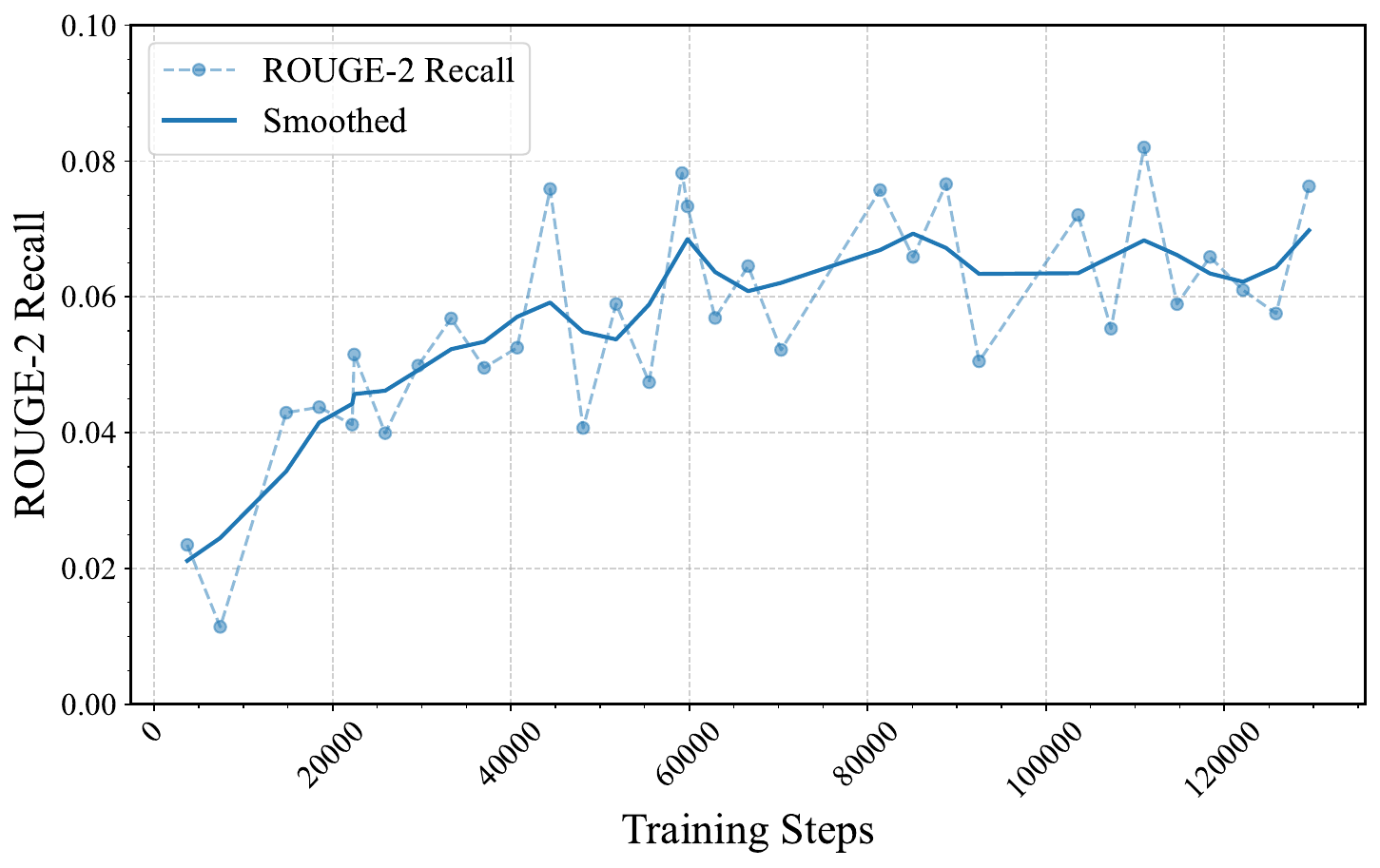}
        \caption{}
        \label{fig:rouge2recall}
    \end{subfigure}
        
    \begin{subfigure}{\columnwidth}
        \centering
        \includegraphics[width=\columnwidth]{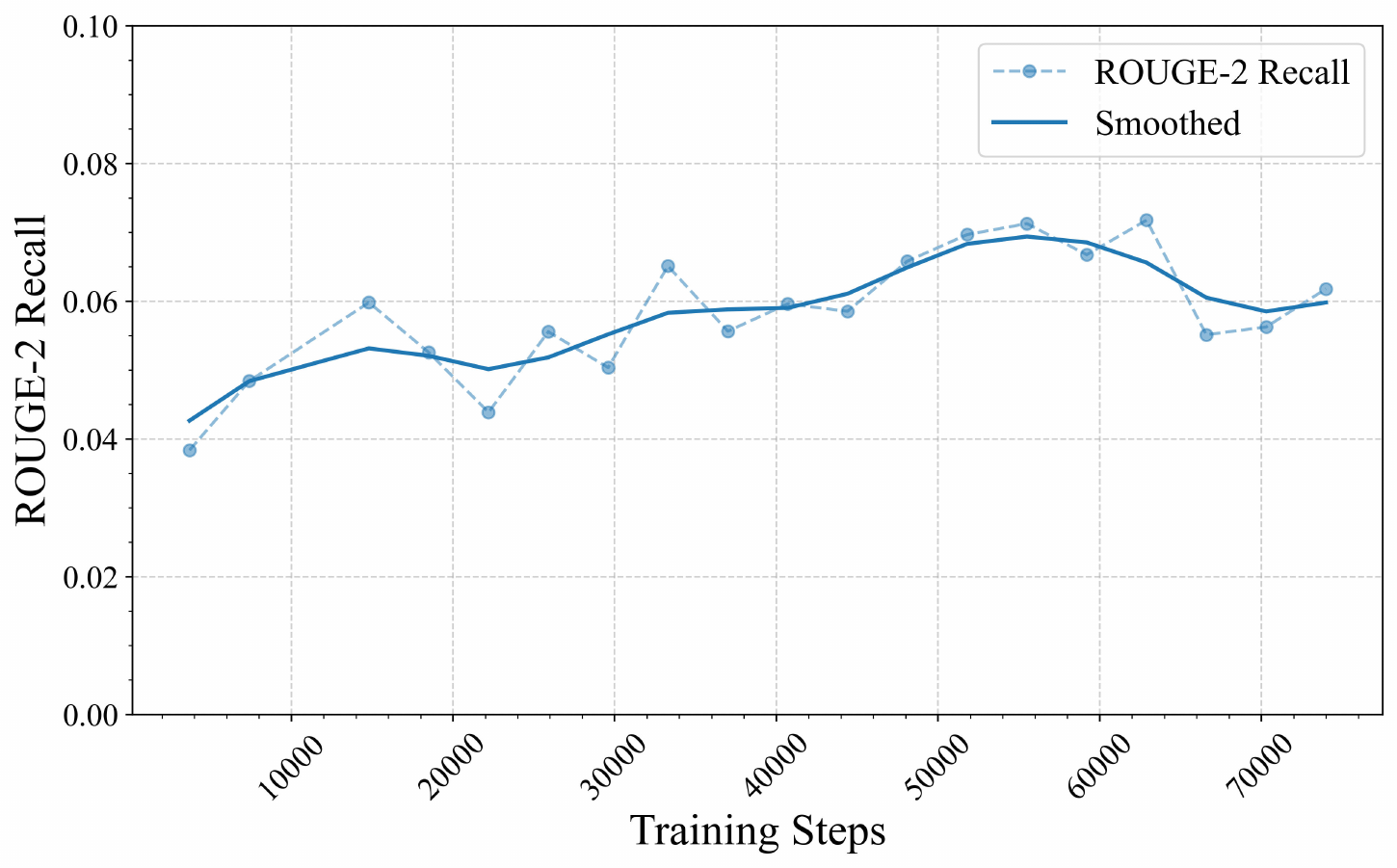}
        \caption{}
        \label{fig:rouge2recall_llama}
    \end{subfigure}
    
    \caption{Reconstruction attack results of Rouge-2 recall values for documents reconstructed at multiple checkpoints. (a) Results on T5. (b) Results on Llama.}
    \label{fig:rouge2recall-all}
\end{figure}

\begin{figure}[!t]
\centering
\includegraphics[width=\columnwidth]{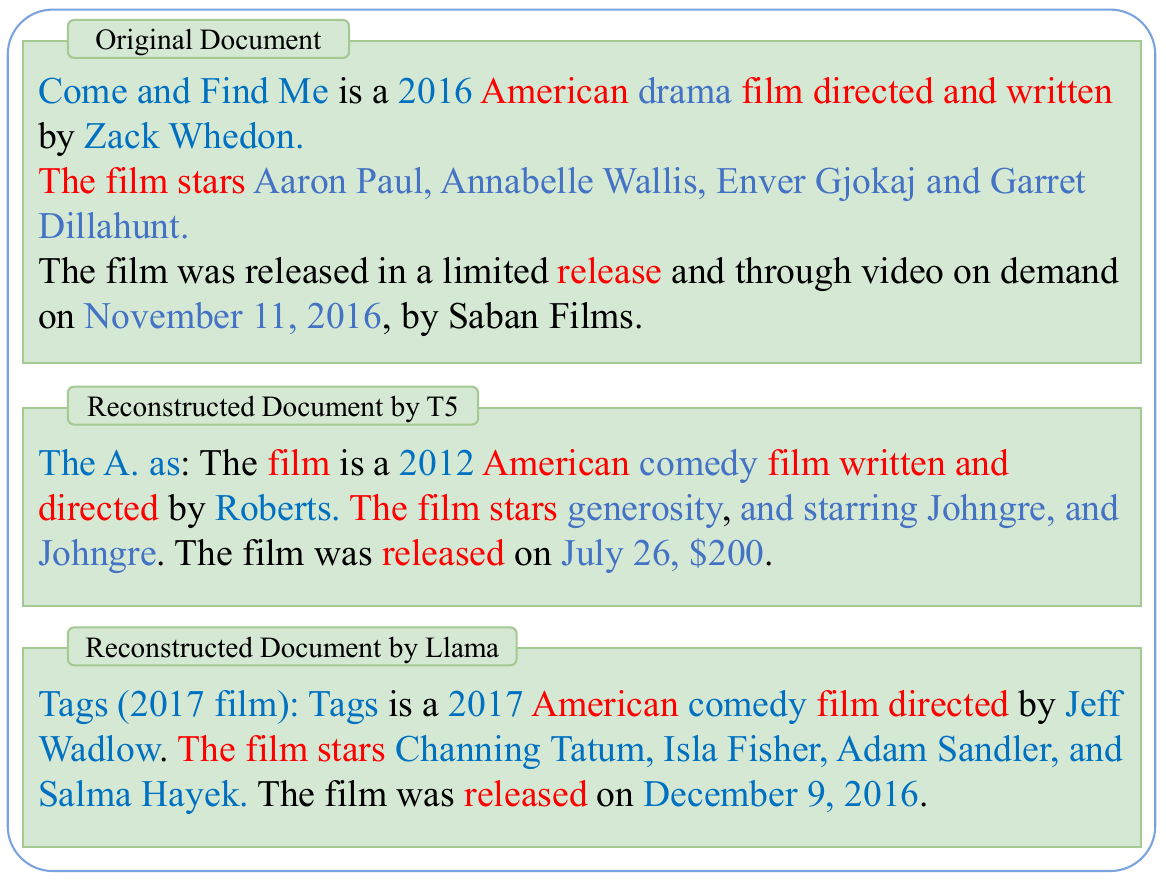}
\caption{\label{fig:rc-case} An example of a document reconstruction. Red indicates successful reconstruction, and blue indicates incorrect reconstruction.}
\end{figure}
Although our approach prevents plaintext exposure of user documents during cloud-based inference, there remains a potential risk that an attacker could attempt to reconstruct documents from generated LoRA weights. We consider a worst-case scenario in which an attacker obtains a set of document-LoRA weight pairs. To simulate such an attack, we use 30,000 documents, selecting 100 as a test set and the rest for training a T5 decoder\footnote{https://huggingface.co/google/long-t5-tglobal-base}~\cite{guo-etal-2022-longt5} to reconstruct documents from LoRA parameters. We use ROUGE-2 recall to measure reconstruction quality. As shown in Figure~\ref{fig:rouge2recall}, even after extensive training (over 10 epochs), the reconstruction recall does not exceed 9\%. 
We further design a stronger reconstruction attack by using Llama-3.2-1B-Instruct as the reconstruction model and feeding it the LoRA parameters, document mask token, and start token for autoregressive decoding. 
As shown in Fig.~\ref{fig:rouge2recall_llama}, the stronger model does not yield a more threatening attack.

Figure~\ref{fig:rc-case} presents an example of document reconstruction. This represents one of the best reconstruction cases observed. In most instances, the T5-based reconstruction consists primarily of repetitive or meaningless phrases such as "Count of the Count of the\ldots". Even in this relatively successful example, key information such as dates, names, and titles is completely incorrect. The Llama-based reconstruction consistently produces fluent text in our observations, but it still fails to recover key information. This demonstrates that our parameter generator is trained to encode abstract knowledge, rather than to directly reconstruct the original text. As a result, recovering the original document content, especially precise private information, from LoRA weights is extremely difficult. This further confirms the reliability of our method. 


\end{document}